\documentclass[conference]{IEEEtran}
\IEEEoverridecommandlockouts
\usepackage{cite}
\usepackage{amsmath,amssymb,amsfonts}
\usepackage{graphicx}
\usepackage{textcomp}
\usepackage{xcolor}
\usepackage{framed,multirow}
\usepackage{booktabs}
\usepackage{algorithm}
\usepackage{algpseudocode}
\usepackage{latexsym}
\usepackage{threeparttable}
\usepackage{tabu}
\usepackage{multicol}
\usepackage{fancyhdr}

\def\BibTeX{{\rm B\kern-.05em{\sc i\kern-.025em b}\kern-.08em
T\kern-.1667em\lower.7ex\hbox{E}\kern-.125emX}}

\fancypagestyle{firstpage} {
\fancyhf{}
\fancyhead[C]{12th International Conference on Computer and Knowledge Engineering (ICCKE 2022), November 17-18, 2022, Ferdowsi University of Mashhad, Iran}
\fancyfoot[L]{978-1-6654-7613-3/22/\$31.00 \copyright 2022 IEEE}

\setlength{\headheight}{22.5pt}
}

\title{Persis: A Persian Font Recognition Pipeline Using Convolutional Neural Networks}

\begin{document}

\renewcommand\IEEEkeywordsname{Keywords}

\author{
    \IEEEauthorblockN{Mehrdad Mohammadian}
    \IEEEauthorblockA{Department of Computer Engineering\\ Islamic Azad University,  Mashhad Branch \\ Mashhad, Iran
    \\ mehrdad.mhmdn@gmail.com} \\
    \and
    \IEEEauthorblockN {
    Neda Maleki, 
    Tobias Olsson, 
    Fredrik Ahlgren
    }
    \IEEEauthorblockA{Department of Computer Science and Media Technology\\ Linnaeus University\\ Kalmar/Växjö, Sweden \\ 
    \{neda.maleki, tobias.olsson, fredrik.ahlgren\}@lnu.se}
}

\maketitle
\thispagestyle{firstpage}

\begin{abstract}
What happens if we encounter a suitable font for our design work but do not know its name? Visual Font Recognition (VFR) systems are used to identify the font typeface in an image. These systems can assist graphic designers in identifying fonts used in images. A VFR system also aids in improving the speed and accuracy of Optical Character Recognition (OCR) systems. In this paper, we introduce the first publicly available datasets in the field of Persian font recognition and employ Convolutional Neural Networks (CNN) to address this problem. 
The results show that the proposed pipeline obtained 78.0\% top-1 accuracy on our new datasets, 89.1\% on the IDPL-PFOD dataset, and 94.5\% on the KAFD dataset. Furthermore, the average time spent in the entire pipeline for one sample of our proposed datasets is 0.54 and 0.017 seconds for CPU and GPU, respectively. We conclude that CNN methods can be used to recognize Persian fonts without the need for additional pre-processing steps such as feature extraction, binarization, normalization, etc.
\end{abstract}


\section{Introduction}
\begin{figure}[t] \centering{
\includegraphics[scale=0.39, clip, trim=0cm 0cm 0cm 0cm]{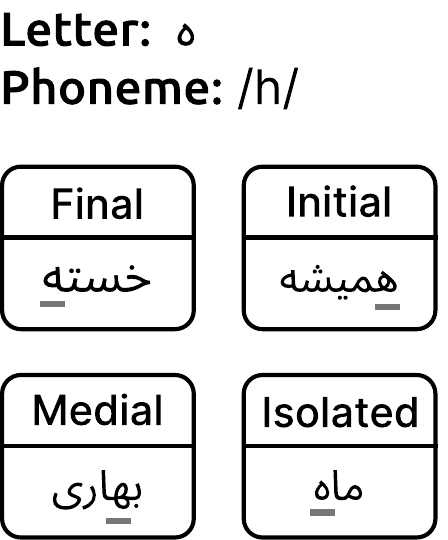}}
\caption{Example of a Persian letter with four different forms of writing.}
\label{fig:letters}
\end{figure}

In 2021, Persian ranked as the fifth most common language found on the web \cite{23}. Consequently, it is of paramount importance to develop new algorithms capable of understanding Persian across various domains, including VFR systems.
OCR is the process of converting handwritten or printed text images into editable and machine-readable text \cite{31}. Recognizing fonts used in images holds fundamental significance in document analysis, serving two primary purposes: OCR and script identification \cite{26, 27}. Font recognition can significantly enhance the speed and accuracy of OCR systems \cite{24, 21}.
The recognition of font typeface, weight, and slope in text images is a content-independent process referred to as VFR \cite{25}. VFR systems find application in assisting graphic designers in identifying and employing fonts encountered in images \cite{22}.
The Persian alphabet comprises thirty-two letters, with some letters featuring additional components, such as dots, dot groups, or slanted bars. Moreover, Persian letters can manifest in final, initial, medial, and isolated forms \cite{34}, as illustrated in Fig. \ref{fig:letters}. In total, there are 114 distinct letter forms within the Persian alphabet \cite{16}.
Both printed and handwritten Persian scripts exhibit a cursive nature \cite{34}. The combination of cursive writing and the unique characteristics of the Persian language renders it a more challenging script to recognize compared to Latin \cite{34}.

The recent papers on Persian font recognition, as presented in Table \ref{tab:datasets}, employ various feature extraction methods but typically focus on handling only one text level (e.g., block, line, word, letter).
These studies often rely on traditional feature engineering techniques such as Histogram of Oriented Gradients (HOG) and Scale Invariant Feature Transform (SIFT) to create features from raw input samples \cite{36}. These methods are designed to construct features directly from the input data \cite{36}.
For feature extraction from images, the Gabor filter and its derivatives, such as Directional Elliptic Gabor (DEG), are commonly used in the literature \cite{32}.
To perform classification, these handcrafted features are then input into various machine learning algorithms. Examples of frequently employed algorithms include Support Vector Machines (SVM), K-Nearest Neighbors (KNN), Multi-Layer Perceptron (MLP), and Random Forest \cite{7, 11}.
However, it is worth noting that the datasets used in recent studies often exhibit several limitations. The samples tend to be similar and do not encompass challenging conditions such as varying backgrounds, text colors, and text positions. Additionally, the number of supported fonts investigated in these datasets is often insufficient for comprehensive font recognition tasks.

In this paper, we aim to address the limitations present in previously proposed datasets for Persian font recognition. To achieve this goal, we introduce two novel datasets, which feature sixty common and freely available Persian fonts. Our datasets encompass samples at four different text levels: block, line, word, and letter. To enhance the realism of our datasets, we introduce variability in text color, text position, and background. Furthermore, we apply four distinct effects to the images, including gradient light, folding, subtle noise, and ink bleed.
As the first publicly available datasets specifically tailored for Persian font recognition, our proposed datasets are accessible through \cite{41}.
Additionally, we present a comprehensive pipeline that leverages CNNs to address the Persian font recognition challenge. CNN models have consistently demonstrated exceptional performance in various computer vision tasks since the early 2000s \cite{28}.
It is worth noting that, in contrast to recent studies on Persian font recognition, which have not explored the use of CNN models, our approach aligns with practices employed in studies focused on languages like English and Arabic. In this paper, we report the results of four distinct experiments conducted within our pipeline, with detailed findings provided in Section \ref{section:experiments}.
\section{Background} 
VFR refers to the process of recognizing the font typeface within an image in a content-independent manner \cite{25}.
Persian font recognition is employed to identify fonts, particularly in the context of the Persian language \cite{7}.
In the most recent papers on Persian font recognition, researchers have introduced methods by engineering and designing feature extractors \cite{8}. These feature extractors transform the original images into feature vectors, and often, a classifier is subsequently trained to categorize the inputs into the desired classes \cite{28}.
Traditional machine learning techniques exhibit limited capabilities in processing raw data \cite{28}. As a result, recent methods in Persian font recognition typically require a series of pre-processing steps, including feature extraction, binarization, and normalization, to prepare the data \cite{1,2}.

\subsection{Deep Learning}
In recent decades, machine learning has gained immense popularity across various research domains, including image classification, recommendation systems, text mining, and more \cite{36}. Among the well-established machine learning algorithms, deep learning stands out as one of the most widely utilized approaches \cite{36}.
Deep learning methods are grounded in representation learning, which enables them to autonomously discover features without manual engineering, making them highly applicable in classification and detection tasks \cite{28}.
In the realm of deep learning applications, such as computer vision, natural language processing, and speech recognition, CNNs have emerged as popular and extensively employed tools \cite{36}. CNNs draw inspiration from neurons and the visual cortex section of both human and animal brains \cite{36}.
In the context of this study, we employ methods rooted in Deep CNN models (DCNN). The term 'deep' signifies the presence of multiple hidden layers within the neural network architecture \cite{36}. Over recent decades, DCNNs have achieved remarkable success in tasks involving segmentation, classification, and detection within the field of computer vision \cite{28}.

\subsection{Image Segmentation}
In the context of font recognition tasks, text present in images can exhibit various backgrounds, ranging from plain paper and textured surfaces to even natural scenes.
Image segmentation has been a challenging problem in the field of computer vision for the past three decades \cite{36}. The primary goal of image segmentation is to make images more simple and meaningful \cite{38}. It serves to enhance image smoothness and facilitates easier evaluation. Additionally, image segmentation aids in identifying regions of interest within an image \cite{38}.
One prominent application of image segmentation is the partitioning of an image into distinct segments, with background segmentation being one of the most popular use cases \cite{33}.
CNNs have proven to be highly effective in image segmentation tasks, with architectures like U-Net standing out. U-Net achieved remarkable success in the 2015 ISBI cell tracking challenge and secured a substantial victory \cite{15}. U-Net is built upon encoder-decoder network structures \cite{33}, which extensively leverage CNN architectures in the domain of deep learning \cite{37}.
Image segmentation plays a crucial role in enabling font recognition in images, as it allows for background-independent font recognition, ensuring that the surrounding scene does not impact the recognition process.

\subsection{Image Classification}
The central step in the font recognition task involves the classification of images into specific font typeface classes. As mentioned earlier, CNN architectures are widely employed in image classification tasks. These typical CNN models consist of a series of convolutional layers followed by pooling layers, often utilizing the max pool layer \cite{36}.
In the final layers of such models, fully connected layers or similar MLPs are commonly utilized \cite{36}. Fully connected layers play a crucial role in transforming multi-dimensional feature maps into a 1-dimensional feature vector \cite{43}.
For instance, in \cite{19}, the first fully connected layer known as FC1 contains 4096 nodes and connects to the 9216 nodes of the output from the preceding pooling layer, MaxPool3. Consequently, there are 37 million weight parameters between these two layers. The high number of parameters in such configurations demands substantial memory and computational resources \cite{43}.
To address this issue, Global Average Pooling (GAP) layers are often used in place of fully connected layers to reduce the overall model parameters \cite{43}. Unlike fully connected layers, GAP layers have no parameters to optimize \cite{43}.
The GAP layer applies a global average pooling operation to each input feature map, effectively summing out spatial information. This design choice enhances the model's robustness to spatial transformations in the input samples \cite{43}.
\section{Related Work}

Font recognition involves the task of identifying the specific font typeface used within an image. In this section, we conduct a review of the most recent papers published on the subject of font recognition, particularly focusing on the Persian language.
In recent papers, Gabor filters have been a prominent choice for feature extraction from images. The following are some notable applications of Gabor filters in font recognition.
In \cite{1}, Gabor filters were applied to normalized versions of images. Their normalization process involved binarization using the Otsu algorithm and the computation of horizontal and vertical projection profiles.
In \cite{42}, a method was proposed for English font recognition that utilized Gabor filters for feature extraction. The Otsu algorithm was employed to reduce grayscale images to binary form. Subsequently, a SVM classifier was trained using the extracted feature vectors.
In \cite{10}, a novel filter named DEG, derived from basic Gabor filters, was introduced. Features extracted by DEG filters are particularly effective in describing the straightness and curvature of text in images, contributing to improved recognition accuracy.
In \cite{11}, two distinct methods leveraging Gabor filters were proposed. The first method, known as Gabor-PCA-MLP, involved feature extraction with Gabor filters, followed by Principal Component Analysis (PCA) to reduce feature dimensions. Then, an MLP was employed for classification. In the second method, a random forest served as the classifier.

While Gabor filters are effective for feature extraction, one of their main drawbacks is their computational time consumption \cite{2}. To address this issue, several alternative feature extraction methods have been proposed in recent research.
In \cite{2}, a new feature extractor called Sobel–Roberts Features (SRF) was introduced as a computationally more efficient alternative to Gabor filters. Their method achieved an accuracy of 94.16\% on a dataset featuring 10 Persian fonts.
In \cite{8}, a method was proposed that combines features from SRF and the wavelet transform to reduce errors. The classification was performed using an MLP, resulting in an accuracy of 95.56\%.
Another approach utilizing the feature combination technique is presented in \cite{3}. This method is based on fractal geometry and involves combining features derived from the Box Counting Dimension (BCD), Dilation Counting Dimension, and Diffusion Limited Aggregates (DLA). BCD and DLA are applied to the binarized versions of images using Niblack's method. Classification is performed using Radial Basis Function (RBF) and KNN classifiers, with reported accuracy rates of 96\% and 91\%, respectively.

Feature descriptors represent another widely adopted approach in font recognition literature.
In \cite{5}, SIFT descriptors are employed. SIFT features are renowned for their ability to enhance the robustness of a method against variations in size, scale, and rotation. This approach eliminates the need for extensive pre-processing steps and reports an impressive accuracy rate, nearly reaching 100\%, when applied to 20 Persian fonts.
In \cite{12}, an improved version of Speeded Up Robust Features (SURF) is presented. The method incorporates the Redundant Key Point Elimination technique to enhance the matching step of the SURF algorithm. Additionally, the nearest-neighbor distance ratio is applied to improve the matching process.

Font recognition can be applied to different text levels, such as word, line, letter, block, and more. For instance, in \cite{8}, all image samples within their dataset are at the line level, meaning that each image consists of a single line of text. In the context of Persian font recognition, the task is also carried out at the letter level.
In \cite{6}, a method based on the Euclidean distance between spatial descriptors and gradient values at boundary points of letters was introduced. One notable advantage of this approach is its robustness against noise, and it requires only a few letters from a document image to recognize the font.
Another approach, detailed in \cite{4}, aims to recognize both the font typeface and font size of an image. This method employs two types of features: the horizontal projection profile and the bounding box of holes within letters. Additionally, binarization is applied to images as a pre-processing step for both types of features.

In \cite{7}, a semi-supervised learning method is introduced for font recognition. This method incorporates a self-training approach that effectively classifies both labeled and unlabeled data.
Within their approach, a majority vote technique based on three algorithms (SVM, RBF Neural Network, and KNN) is employed to partition unlabeled data into reliable and unreliable segments.
In \cite{9}, an algorithm independent of content is proposed for font recognition. The algorithm relies on the identification of most-frequent connected components. Additionally, they introduce a voting algorithm to facilitate font recognition in images.

All the reviewed papers on Persian font recognition share certain limitations in their utilized datasets. These limitations include a low number of font types represented in the datasets and uniformity in the properties of the samples, such as text color, text positioning, background, and lighting conditions. Such uniformity does not adequately reflect the diverse real-world scenarios where fonts are encountered. Additionally, these datasets typically do not support the simultaneous presence of multiple text levels.
Furthermore, in most of the reviewed Persian font recognition papers, the proposed methods require additional computational steps for preprocessing tasks such as feature extraction, binarization, normalization, noise removal, dot deletion, and horizontal and vertical projection profiling, among others.
In contrast, our proposed pipeline stands out by eliminating the need for these preprocessing steps on images.
\section{Proposed Datasets}
\begin{figure}[h] \centering{
\includegraphics[scale=0.205, clip, trim=0cm 0cm 0cm 0cm]{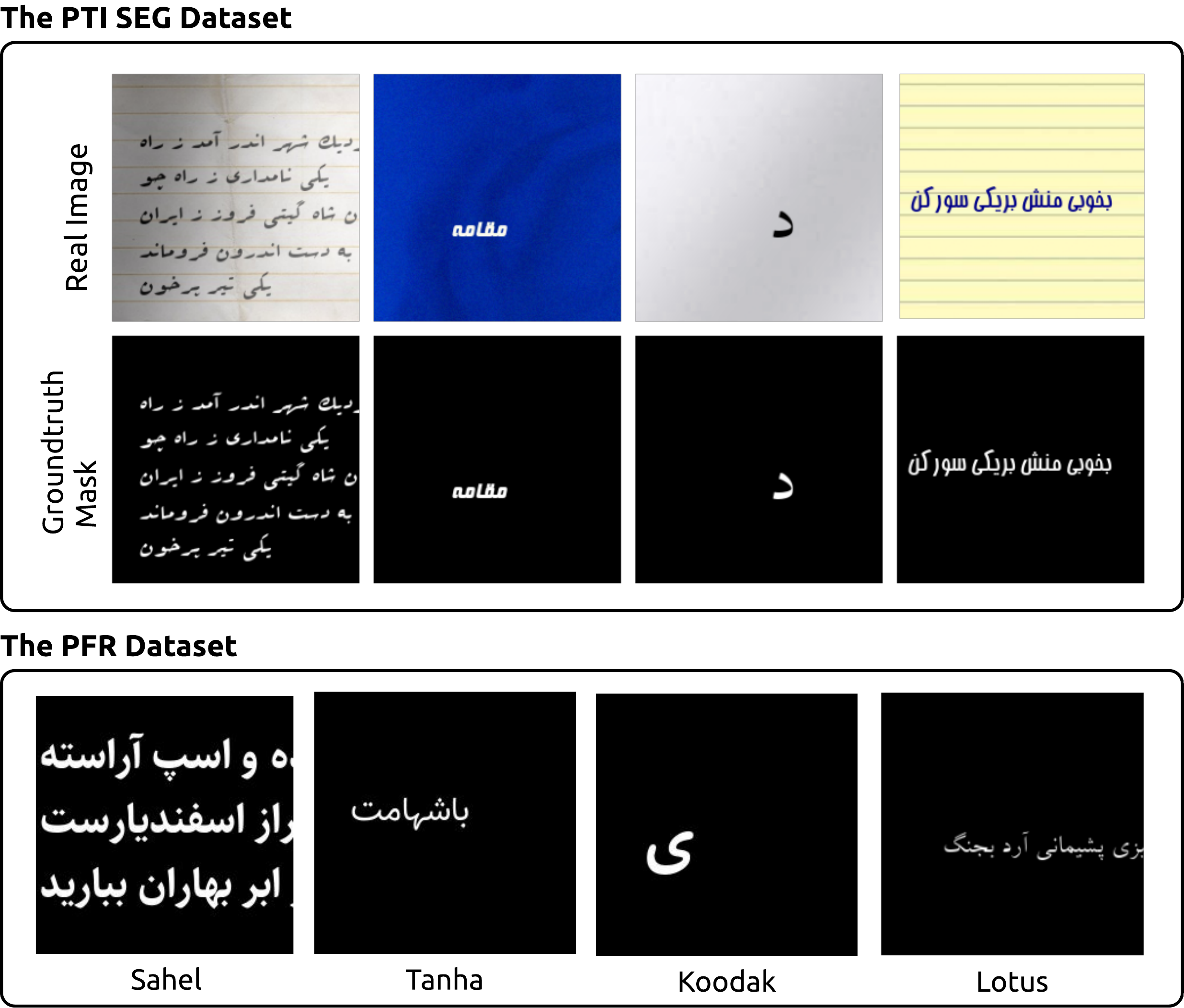}}
\caption{Some samples of the PFR and PTI SEG datasets.}
\label{fig:samples}
\end{figure}

The initial step in addressing a font recognition task involves the collection of labeled images. Notably, the datasets utilized in recent papers focusing on Persian font recognition, which we have reviewed in preceding sections, exhibit various limitations when it comes to constructing a robust and versatile font recognition system. For instance, previous datasets lack diversity in terms of factors such as lighting conditions, text colors, text positioning, and backgrounds.
To bridge this significant gap in existing datasets and introduce more challenging samples for Persian font recognition, we have introduced the Persian Font Recognition (PFR) and Persian Text Image Segmentation (PTI SEG) datasets. Furthermore, the real image segment of the PTI SEG dataset can be effectively employed for testing various VFR systems.
In Table \ref{tab:datasets}, we provide a comparative analysis of our datasets alongside other existing datasets. Additionally, we offer a visual representation of some sample images from our datasets in Fig. \ref{fig:samples}.
As demonstrated in Table \ref{tab:datasets}, all samples in recent datasets feature black text on a white background, and the positioning of text within the images is highly uniform.

\begin{figure}[h] \centering{
\includegraphics[scale=0.2]{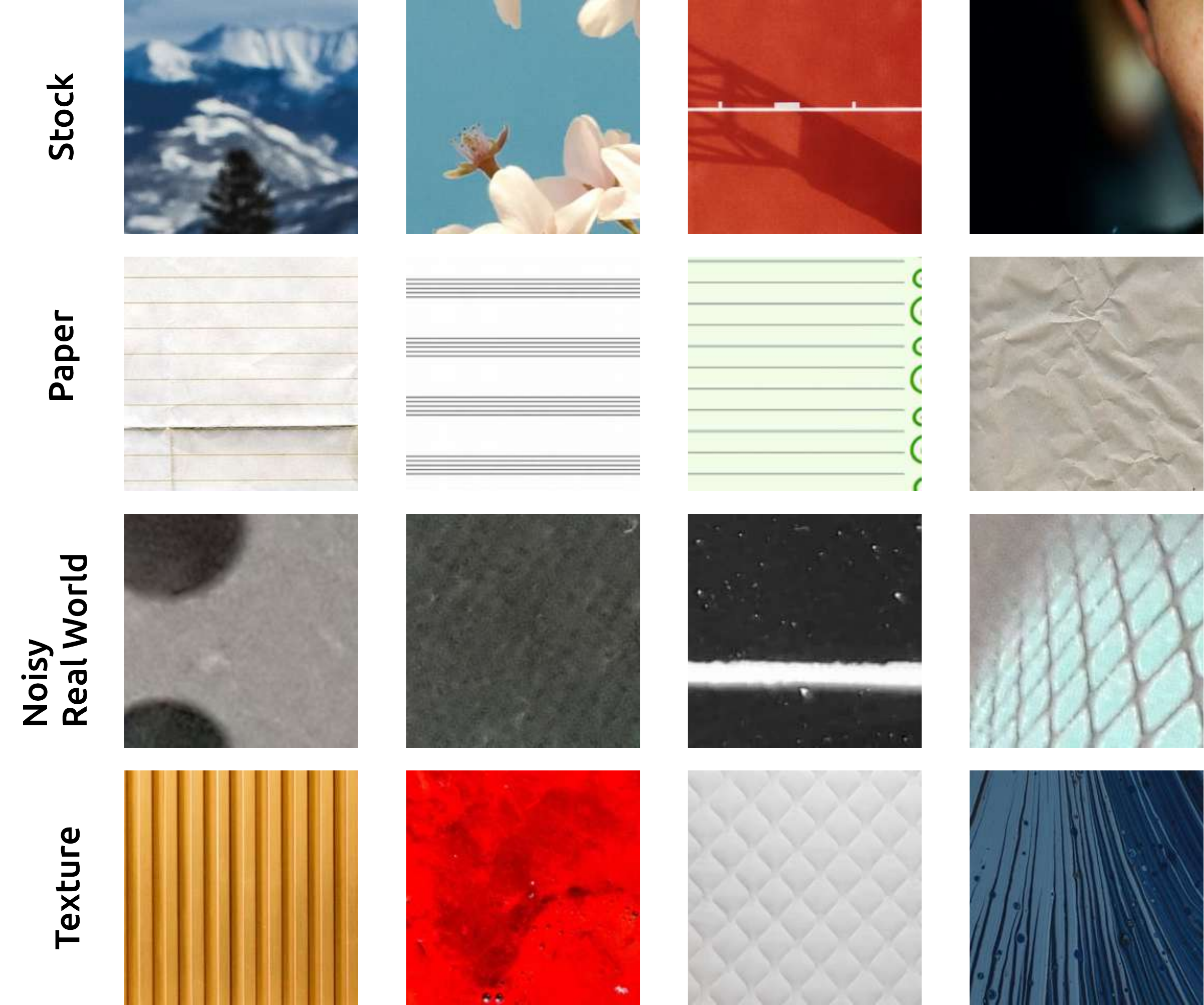}}
\caption{Examples of different types of backgrounds that we used in the PTI SEG dataset.}
\label{fig:backgrounds}
\end{figure}

\begin{figure}[t] \label{fig4} \centering{
\includegraphics[scale=0.215]{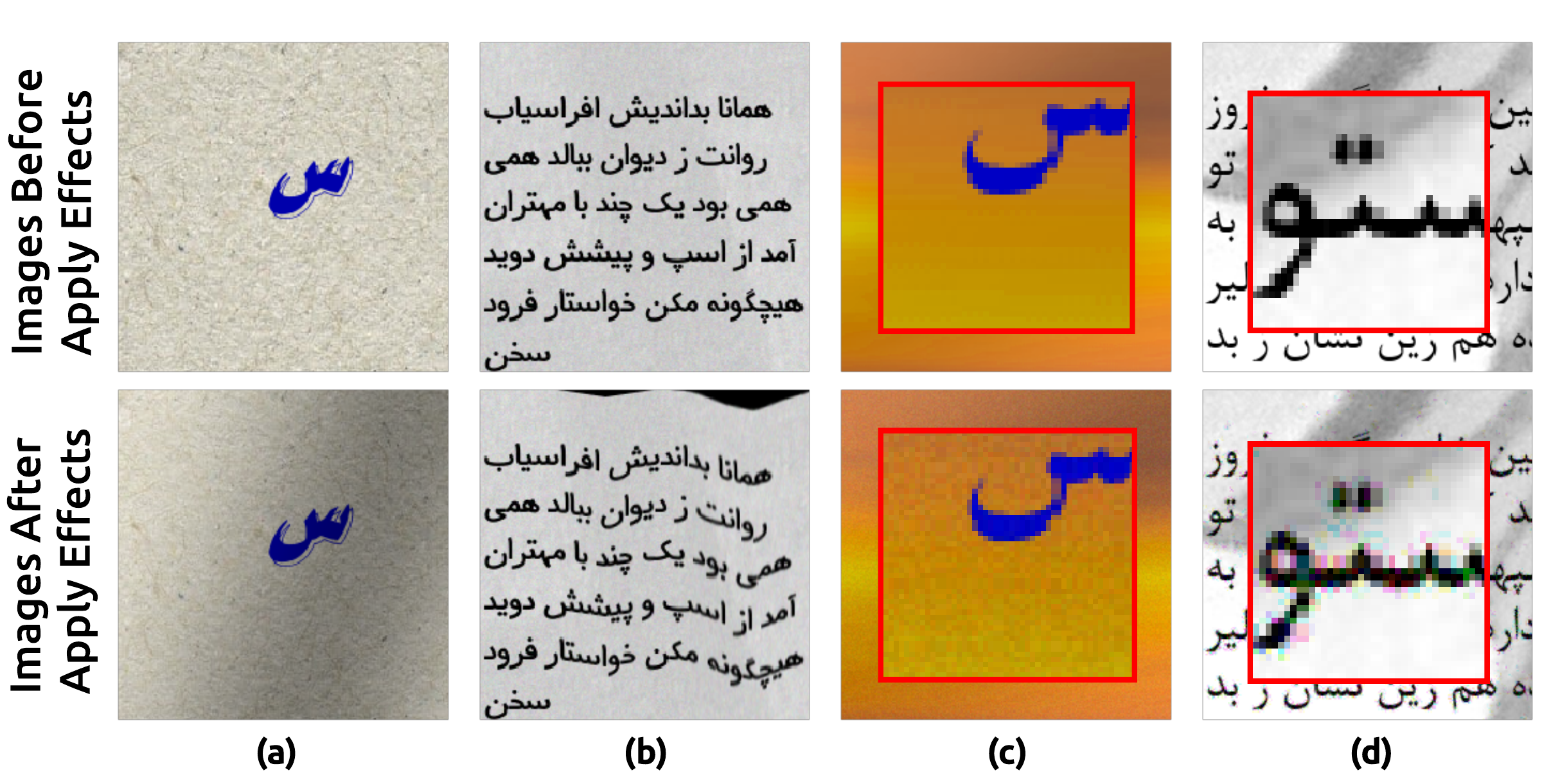}}
\caption{Examples of four types of effects that we applied to samples of the PTI SEG dataset - (a): Gradient Light, (b): Folding, (c): Subtle Noise, (d): Ink Bleed.}
\label{fig:effects}
\end{figure}

\begin{table*}[ht]
\centering
\caption{Comparison of previous datasets with our proposed datasets.}
\label{tab:datasets}
\begin{tabular*}{\linewidth}{l@{\extracolsep{\fill}}*{8}{c}}

\toprule
\textbf{Paper}\tnote{*} & \textbf{Fonts} & \textbf{Text level} & \textbf{Text color} & \textbf{Background}  & \textbf{Samples}  \\
\midrule
\cite{1} & 5 &  block  &  black  &  white & - \\
\cite{2} & 10 &  line  &  black  &  white & 20000 \\
\cite{3} & 10 &  word  &  black  &  white & 2000 \\
\cite{4} & 7 &  document  &  black  &  white & 845 \\
\cite{5} & 20 &  document  &  black  &  white & 1495 \\
\cite{6} & 25 &  letter  &  black  &  white & 500 \\
\cite{8} & 10 &  line  &  black  &  white & 2100 \\
\cite{10} & 10 &  line  &  black  &  white & 600 \\
The PTI SEG dataset & 60 &  block, line, word, letter  &  random  & random  & 10000 \\
The PFR dataset & 60 &  block, line, word, letter  &  white  & black  & 20000 \\

\bottomrule
\end{tabular*}
\end{table*}
In the generation of the PTI SEG dataset, we employ a total of 735 distinct backgrounds, categorized into four types: stock backgrounds, paper textures, backgrounds simulating noisy real-world conditions, and textured backgrounds. Additionally, we apply four distinct effects to the images, as illustrated in Fig. \ref{fig:backgrounds} and Fig. \ref{fig:effects}. These effects include gradient lighting, folding, subtle noise, and ink bleed.

As is known, real-world images can vary significantly in terms of lighting conditions. To account for this variability, we have applied a random gradient lighting effect to our images for simulation. Furthermore, considering that many real-world images may exhibit noise and ink bleed around the text, we have incorporated subtle noise and ink bleed effects to replicate such scenarios. Additionally, some images may resemble folded book pages, prompting us to apply a folding effect to capture this characteristic.
To obtain stock and textured images, we utilized the Unsplash API \cite{30}. The noisy real-world images are photographs taken by one of the authors, while paper images were collected through Google image search under a Creative Commons license.

Both the PFR and PTI SEG datasets encompass four text levels (block, line, word, and letter) and feature sixty different font typefaces. The number of samples for each class of Persian fonts is uniform across all classes. The image dimensions in both datasets are set to 224$\times$224 pixels. The real images in the PTI SEG dataset are in RGB color mode, whereas the mask images are in grayscale.
For block and line levels, we utilized poetry from the Shahnameh book dataset in \cite{13}, which consists of nearly 100,000 lines. For word-level data, we employed the Persian word dictionary dataset from \cite{14}. Lastly, for letter-level data, we utilized the thirty-two letters of the Persian language alphabet. In all cases, text and colors are randomly selected and overlaid onto random background images, after which a random effect is applied.
Our dataset generation process is detailed in Algorithm \ref{alg1}.

\begin{algorithm}
\caption{Our data generator algorithm.}\label{alg1}
\begin{algorithmic}[1]
  \State $NumFonts \gets 60$ 
  \State $ datasetSize \gets 10000 $ 
  \State $ eachFontSamples \gets$ datasetSize divided by NumFonts
  \State $ PFRdataset \gets $ True 
  \For {e in range(0, eachFontSamples)}
        \For {font in FontsDir}
            \State $ textLevel \gets$ random choice [1,2,3,4]
            \State $ text \gets$ choice random text from selected textLevel
            \State $ textCoordinate \gets$ generate random x, y
            
            \State $ mskTextColor \gets$ \#FFFFFF
            \State $ mskBackgr \gets $ black background
            \State $ mask \gets$ generate mask with text, mskTextColor, \\\hspace{7em}textCoordinate and mskBackgr     
            
            \If {PFRdataset}
                \State $label \gets$ name of font
                \State Save mask and label
            \Else
                \State $ textColor \gets$ generate random color
                \State $ typeBackgr \gets$ random choice [1,2,3,4] 
                \State $ imgBackgr \gets $ choice random image from \\\hspace{11.1em}selected typeBackgr   
                \State $ image \gets$ generate image with text, textColor, \\\hspace{9em}textCoordinate and imgBackgr
                \State  effect $\gets $ random choice [True, False]
                \If {effect}
                    \State $ image \gets$ apply a random effect on image
                \EndIf
                \State Save image and mask
            \EndIf
        \EndFor
  \EndFor
\end{algorithmic}
\end{algorithm}
\section{Proposed Method}
\begin{figure}[h] \centering{
\includegraphics[scale=0.2]{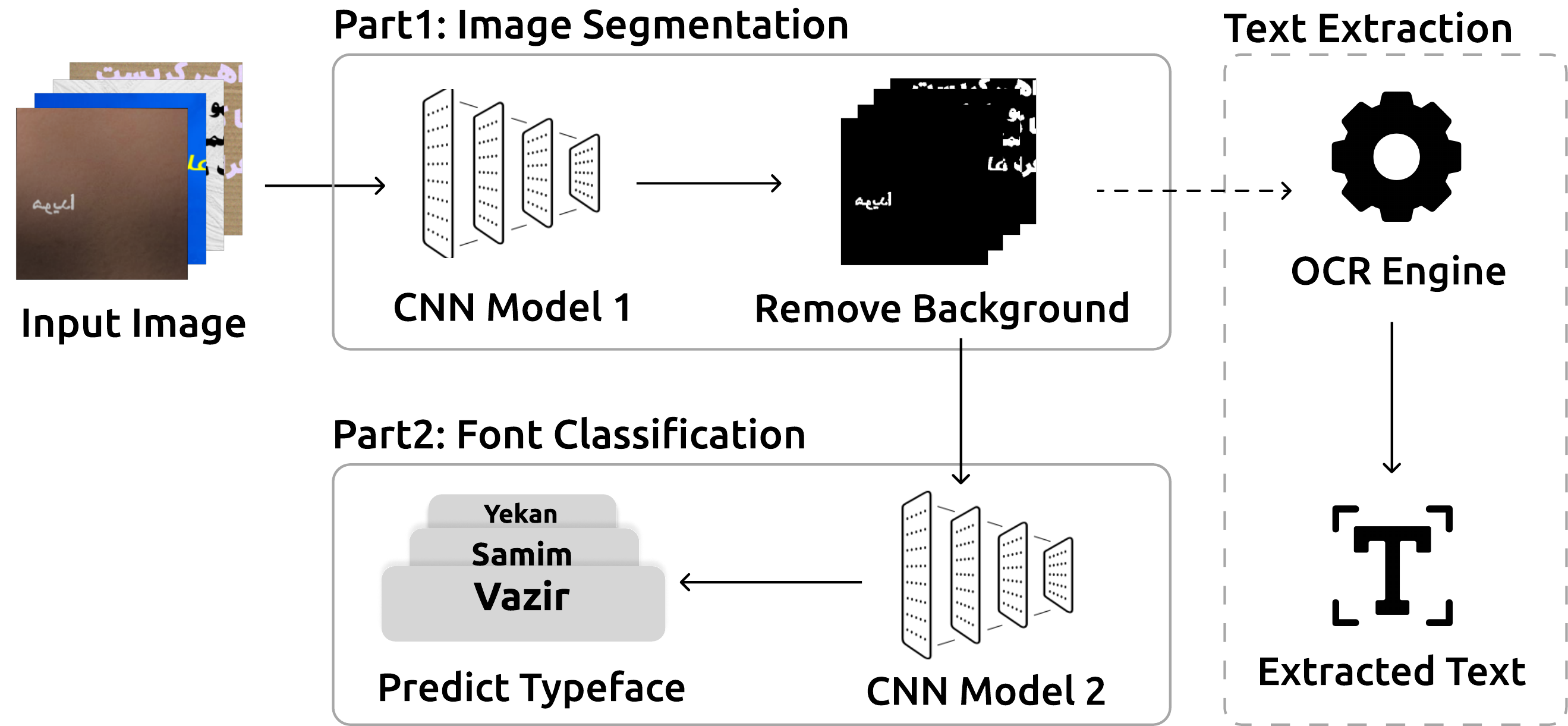}}
\caption{Relationships between different parts of our proposed pipeline and their inputs/outputs.}
\label{fig:block-diagram}
\end{figure}

After preparing the data for addressing the Persian font recognition task, the next step is to train our models. As previously discussed in earlier sections, our approach to Persian font recognition is based on CNN models, comprising two key components. The block diagram of our pipeline is presented in Fig. \ref{fig:block-diagram}.
In the first part, we employ a CNN-based image segmentation model to remove backgrounds from the images. Subsequently, in the second part, another CNN model is utilized to predict the font typeface class depicted in the images. As depicted in Fig. \ref{fig:block-diagram}, the output from the first part can also be integrated with an OCR engine's text extractor. Notably, significant research efforts, such as those detailed in \cite{16}, have already been undertaken in the realm of Persian OCR systems.
In the upcoming subsections, we will provide detailed explanations of each component within our pipeline.

\subsection{Image Segmentation Part}
\begin{figure*}[t] \centering{
\includegraphics[scale=0.385]{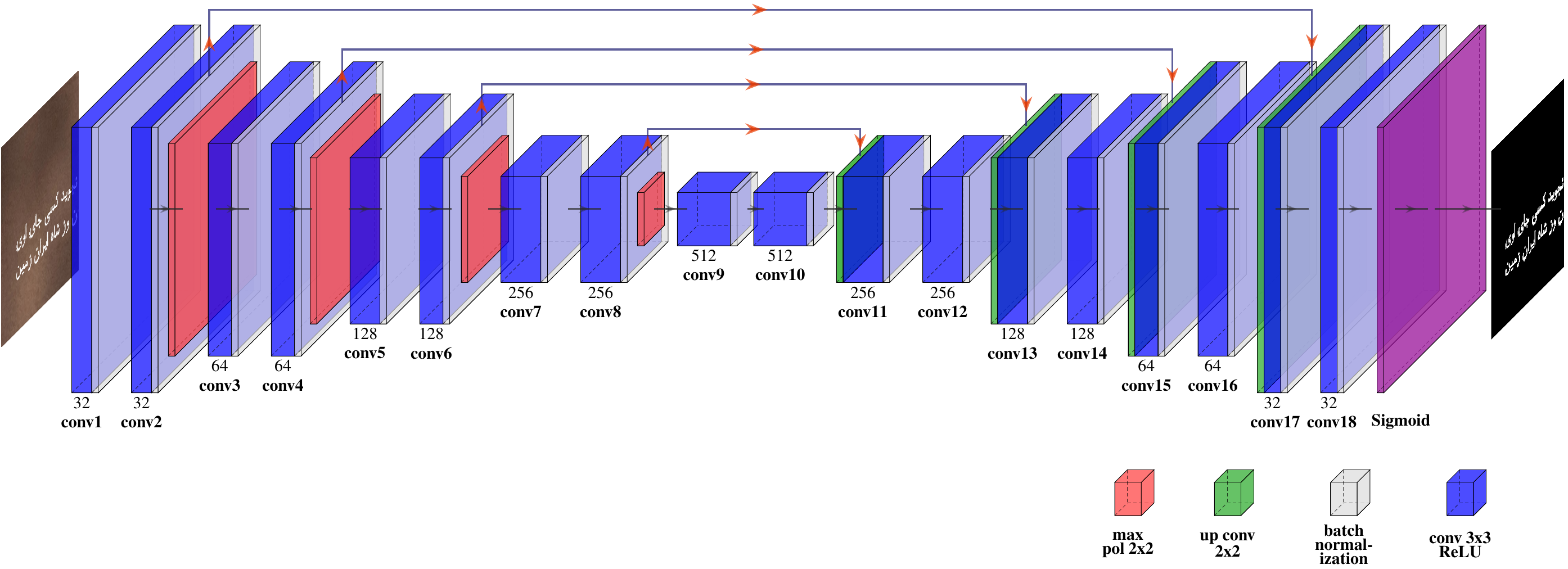}}
\caption{The architecture of the image segmentation model includes the relationship of layers, the number of output filters in the convolution layers, and the type of other layers.}
\label{fig:unet}
\end{figure*}

In this subsection, we delve into the image segmentation component of our proposed pipeline. We have adopted a U-Net-shaped model, characterized as an encoder-decoder CNN. The U-Net model, known for its speed and accuracy in image segmentation, notably achieved a significant victory in the ISBI cell tracking challenge of 2015, surpassing other competing models by a substantial margin \cite{15}. The architectural layout of our segmentation model is illustrated in Fig. \ref{fig:unet}.
As depicted in Fig. \ref{fig:unet}, all convolutional layers employ a 3x3 kernel and utilize the ReLU activation function. Additionally, we incorporate batch normalization layers following each convolutional layer. The red arrows indicate concatenation points between the batch normalization and up-convolution layers.

In this part of the pipeline, we encounter a binary image segmentation challenge characterized by two classes: the background in black and the text in white. Consequently, for the purpose of loss calculation, we employ the binary cross-entropy loss function. In this context, $m$ represents the number of outputs, $y$ signifies the ground truth mask, and $\hat{y_i}$ denotes the predicted mask generated by the model.

\begin{equation}\label{eq1}
Loss = - \frac{1}{m} \sum_{i=1} ^ m y_i \cdot  log \hat {y_{i}} + (1 - y_i) \cdot log(1 - \hat{y_i})
\end{equation}
\\
For evaluation purposes, we employ the Intersection over Union (IoU) metric with thresholds of 50 and 75, along with the Dice similarity coefficient. In this context, $A$ denotes the ground truth mask, and $B$ represents the predicted mask.
\begin{equation}\label{eq2}
IoU = \frac{|A \cap B|}{|A \cup B|}
\end{equation}

\begin{equation}\label{eq3}
Dice = 2 \times \frac{|A \cap B|} {|A| + |B|}
\end{equation}
\\
For the optimization of weights in our proposed image segmentation model, we utilize the Adam optimizer, as introduced in \cite{17}. Adam is known for its computational efficiency and minimal memory requirements \cite{17}.
The learning rate for the optimizer is adjusted manually during training, following a similar approach as in \cite{19}. Initially, the optimizer starts with a learning rate of 1e-5. As we monitor the plateau pattern in validation metrics, we decrease the learning rate to 1e-4.
Regarding model training, we partition the PTI SEG dataset, consisting of 10,000 samples, into three subsets. During training, 80\% of the PTI SEG samples are allocated to the training set, while the remaining 20\% form the test set. Furthermore, 20\% of the training set is set aside for the validation set.

\begin{figure}[t] \centering{
\includegraphics[scale=0.125]{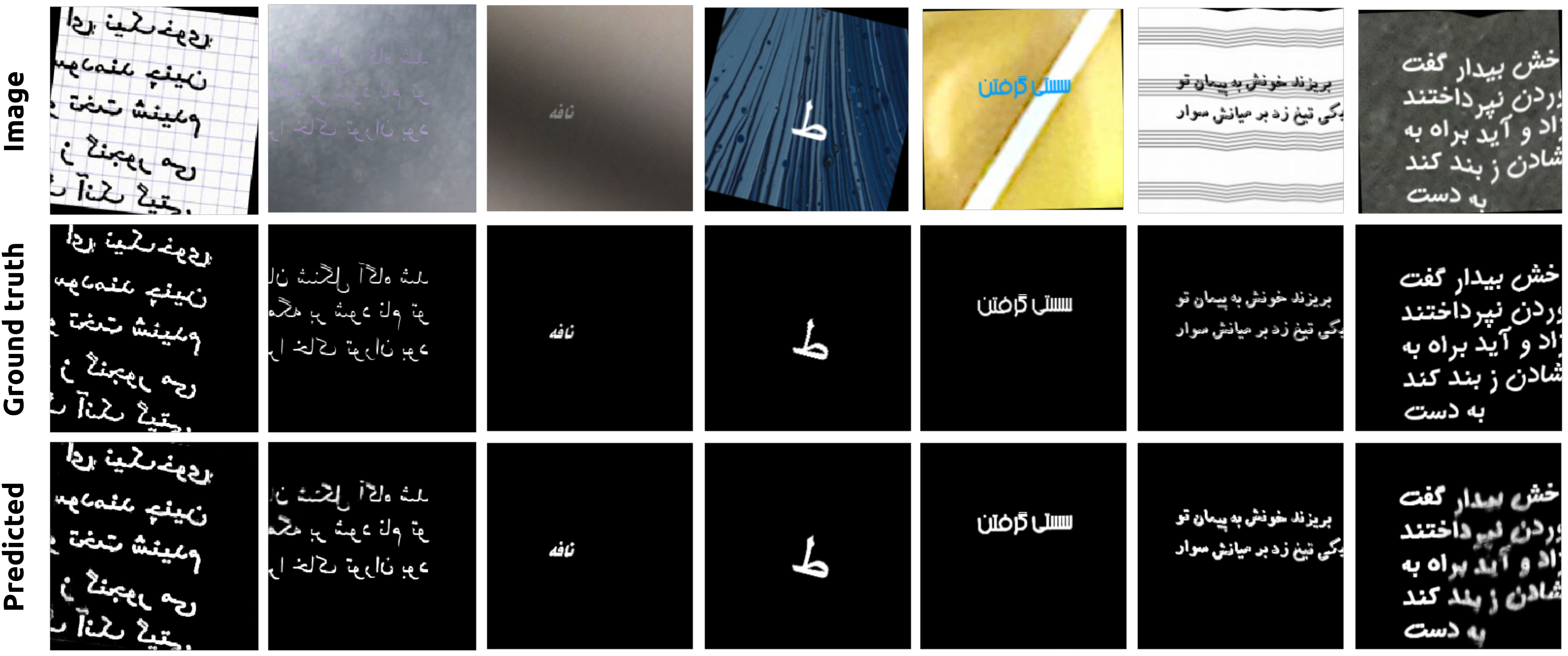}}
\caption{Comparison of predicted masks of the segmentation model and their ground truth.}
\label{fig:unet-results}
\end{figure}

To mitigate the risk of overfitting, we employ three data augmentation techniques during training: vertical and horizontal flips with a 0.3 probability, rotation with a 0.3 probability, and a 30-degree rotation limit applied simultaneously to both images and their masks. These augmentations, along with various random scenarios applied to the images, enhance the robustness of our method against rotation, different types of flips, varying lighting conditions, text colors, text sizes, and backgrounds.
Table \ref{tab:unet-metrics} presents the evaluation metrics results for the test section of the PTI SEG dataset. We report two sets of results: one with random flips and rotations applied to the test set samples and another without these considerations.
In Fig. \ref{fig:unet-results}, we showcase some predicted masks of the images.

\begin{table}[h]
\caption{Experimental results of the image segmentation model on the PTI SEG test set.\label{tab:unet-metrics}}
\centering
\begin{tabular}{cccc}
\toprule
&  \multicolumn{3}{@{}c@{}}{\textbf{Metrics}} \\ \cmidrule{2-4}
\textbf{Augmentation}
& \textbf{IoU$>$50 \%}  & \textbf{IoU$>$75 \%}  & \textbf{Dice \%}  \\
\midrule
considered     &  85.5  &  81.8  &  88.9 \\
not considered &  88.7  &  85.1  &  91.1 \\
\bottomrule
\end{tabular}
\end{table}

\subsection{Image Classification Part}
\begin{figure*}[t] \centering{
\includegraphics[scale=0.48]{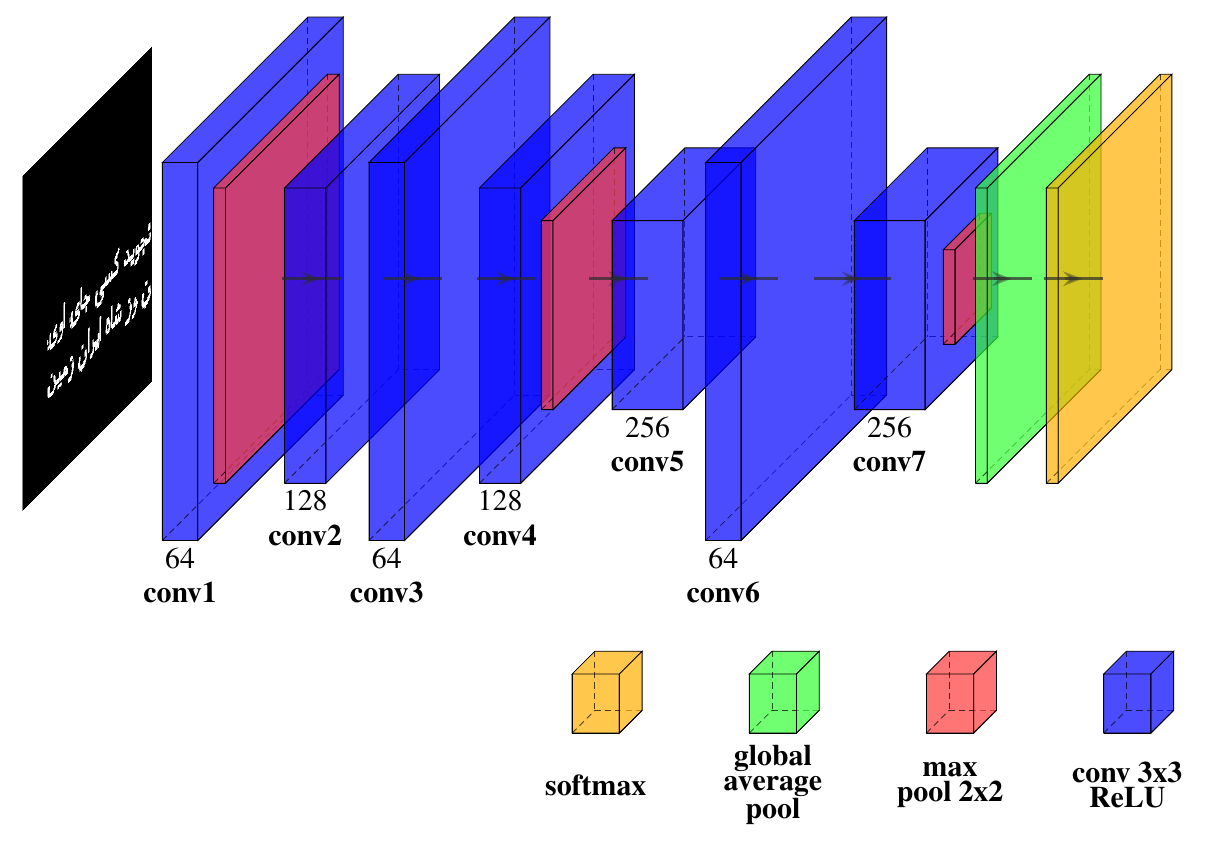}}
\caption{The architecture of the classification model includes the relationship of layers, the number of output filters in the convolution layers, and the type of other layers.}
\label{fig:classification}
\end{figure*}

The final component of our proposed pipeline focuses on image classification. We have developed a compact CNN model with only 827,000 trainable parameters.
In the last layers of our model, we employ GAP, a technique introduced in \cite{18}. By using GAP, we eliminate the need for fully connected layers, which contributes to a reduction in the number of parameters within our model \cite{18}. Moreover, GAP offers the advantage of having no parameters that require optimization, thus helping to mitigate overfitting in this layer \cite{18}.
The architecture of our classification model is depicted in Fig. \ref{fig:classification}. This model takes ground truth masks as inputs and produces the predicted class for the input image as the output.

As illustrated in Fig. \ref{fig:classification}, all convolution layers utilize a 3$\times$3 kernel with the ReLU activation function. In the final layer, we employ the softmax activation function to determine the class of the image.
For the training of our classification model, we utilize the PFR dataset, which consists of 20,000 mask images distributed across 60 classes, featuring diverse and random scenarios similar to those in the PTI SEG dataset. Similar to the image segmentation phase, we partition the PFR dataset to train the classification model. Optimization is carried out using the Adam optimizer with a learning rate set to 1e-4. To mitigate overfitting, we apply data augmentation using the same conditions as in the image segmentation phase.
The evaluation metrics results for the test section of the PFR dataset are presented in Table \ref{tab:metrics-exp1}.

\section{Experimental Results}
\label{section:experiments}
In earlier subsections, we explored various components of our pipeline. In this section, we subject our pipeline to four different tests:
\begin{enumerate}
\item We evaluate the performance of the entire pipeline by testing it on two publicly available datasets designed for Persian OCR and Arabic font recognition.
\item For model testing, we employ grayscale images that closely resemble examples from previously reviewed datasets, as outlined in Table \ref{tab:datasets}.
\item We conduct a comparison between the processing speed of our proposed pipeline and recent studies in the field.
\item Additionally, we analyze the parameters and Floating Point Operations (FLOPs) associated with our pipeline.
\end{enumerate}

\subsection{Experiment 1}
In \cite{20}, the KAFD dataset for Arabic font recognition is introduced. The Arabic and Persian languages share remarkably similar alphabets and writing systems. The KAFD dataset bears a strong resemblance to previously introduced datasets, as shown in Table \ref{tab:datasets}. The Original KAFD is an extensive dataset, comprising 2,576,024 samples, with images available at four different resolution levels, all containing the same content \cite{20}. For our purposes, we have opted to work with a smaller version of the KAFD dataset, which consists of 5,820 samples distributed across 11 classes. All images in this dataset are at the line level and have a resolution of 100 DPI, which represents the lowest quality within the dataset.
In order to ensure the versatility of our pipeline, even with lower-resolution images, we have selected the lowest DPI level. Our image segmentation model, which is trained on the PTI SEG dataset, first removes backgrounds before utilizing the images to train our classification model.

The IDPL-PFOD dataset, proposed for Persian OCR research in \cite{29}, serves as the testing dataset for our pipeline. This dataset comprises 11 classes of images, totaling 30,128 samples, with each class containing nearly 2,740 samples.
The IDPL-PFOD dataset presents more challenging conditions compared to the datasets introduced in Table \ref{tab:datasets}. To enhance the realism of the images, it incorporates various elements such as textured and noisy backgrounds, as well as white backgrounds. Additionally, it applies sloping distortion, sinewave distortion, and blur effects.
After the removal of all backgrounds, we employ these segmented images to train our classification model. The results of this experiment are presented in Table \ref{tab:metrics-exp1}.

\begin{table*}[t!]
\centering
\caption{Speed comparison of the related work and proposed method. \label{tab:speed}}
\begin{tabular}{ccccccc}
\toprule
\textbf{Paper} & \textbf{Method} & \textbf{Processor} & \textbf{Environment}  & \textbf{Feature length} & \textbf{Feature extraction}  & \textbf{Whole process} \\
\midrule

\multirow{5}{*}{\cite{2}} &
16 channels Gabor &  2.4GHz CPU & - &  32   &  0.347 & - \\
& 8 channels Gabor & 2.4GHz CPU & - &  256 &  0.178 & -  \\
& Sobel gradient  & 2.4GHz CPU & - &  256 &  0.0022  & - \\
& Roberts gradient  &  2.4GHz CPU & - &  256  &  0.0018  & - \\
& SRF  & 2.4GHz CPU & - & 512 & 0.00378 & - \\

\multirow{5}{*}{\cite{10}} &
Gabor &  3.4GHz Pentium 4 & MATLAB 7.4 & 96 &  0.468 & - \\
& Sobel-Roberts &  3.4GHz Pentium 4 & MATLAB 7.4 & 512 &  0.023 & - \\
& DEG &  3.4GHz Pentium 4 & MATLAB 7.4 & 102 & 0.481 & - \\
& Gabor \& DEG &  3.4GHz Pentium 4 & MATLAB 7.4 & 198 & 0.949 & - \\
& Sobel-Roberts \& DEG & 3.4GHz Pentium 4 & MATLAB 7.4 & 614 & 0.504 & - \\

\multirow{2}{*}{\cite{10}} &
8 channels Gabor & dual core 2.4GHz Pentium & MATLAB & - & - & 3.3  \\
& Holes of Letters and HPP & dual core 2.4GHz Pentium & MATLAB & - & - &  0.21 \\

\multirow{1}{*}{\cite{8}} &
Sobel-Robert Wavelet & - & - & - & 0.45 & -  \\

\multirow{2}{*}{Our} &
CNN & Intel® Xeon® CPU 2.20GHz & Python 3.7, Linux & - & - & 0.54\\
& CNN & NVIDIA Tesla T4 GPU & Python 3.7, Linux & - & - & 0.017   \\

\bottomrule
\end{tabular}
\end{table*}

\begin{table}[h!]
\caption{Experimental results of the classification model. \label{tab:metrics-exp1}}%
\centering
\scalebox{1}{
\begin{tabular}{ccccc}
\toprule
& & & \multicolumn{1}{@{}c@{}}{\textbf{Metrics}} \\ \cmidrule{3-5}
\textbf{Dataset} & \textbf{Augmentation}\tnote{*}
& \textbf{Top-5 \%}  & \textbf{Top-3 \%}  & \textbf{Top-1 \%}  \\
\midrule

\multirow{2}{*}{PFR} &
considered    &  94.9  &  89.8  &  73.1 \\
& not considered &  96.7  &  92.7  &  78.0 \\
\multirow{2}{*}{IDPL-PFOD} &
 considered    &  97.7  & 93.9  &  79.7 \\
& not considered  &  99.0  &  97.5  &  89.1 \\
\multirow{2}{*}{KAFD} &
considered    &  99.6  &  99.0  &  90.5 \\
& not considered  &  99.8  &  99.8  &  94.5 \\

\bottomrule
\end{tabular}}
\end{table}
\subsection{Experiment 2}
As all samples from recent Persian font recognition datasets are in grayscale, we tested our pipeline on the grayscale version of our new datasets. For this experiment, we trained our image segmentation model on grayscale versions of the PTI SEG dataset samples and then trained our classification model on the output of the image segmentation model. The results of this experiment on the test subset of the PTI SEG dataset are shown in Tables \ref{tab:exp2} and \ref{tab:exp2-2}.

\begin{table}[h]
\centering
\caption{Experimental results of the segmentation model with grayscale images. \label{tab:exp2}}%
\begin{tabular}{cccc}
\toprule
& \multicolumn{3}{@{}c@{}}{\textbf{Metrics}} \\ \cmidrule{2-4}
\textbf{Model}
& \textbf{IoU$>$50 \%}  & \textbf{IoU$>$75 \%}  & \textbf{Dice \%}  \\
\midrule

\multirow{1}{*}{Image Segmentation} &
 80.4  &  79.2  &  58.7 \\  
\bottomrule
\end{tabular}
\end{table}
\begin{table}[h]
\centering
\caption{Experimental results of the classification model with grayscale images. \label{tab:exp2-2}}%
\begin{tabular}{cccc}
\toprule
& \multicolumn{3}{@{}c@{}}{\textbf{Metrics}} \\ \cmidrule{2-4}
\textbf{Model}
& \textbf{Top-5 \%}  & \textbf{Top-3 \%}  & \textbf{Top-1 \%}  \\
\midrule
\multirow{1}{*}{Image Classification} &
86.5  &  80.9  &  65.2 \\  
\bottomrule
\end{tabular}
\end{table}

\subsection{Experiment 3}
We compare the time performance of our proposed pipeline with recent studies. However, most recent studies report time only for the feature extraction steps, while our pipeline based on CNN models is free of these steps, making direct comparisons difficult.

In addition to time, we report the processors and environments used in the recent articles. All reported times are for one sample, and both the feature extraction and whole process columns are in seconds. We report the mean time of 100 predictions of our pipeline on one sample, both on GPU and CPU. The results of this experiment are shown in Table \ref{tab:speed}.
\subsection{Experiment 4}
To enable meaningful comparisons with future studies, we provide essential metrics for our pipeline, specifically the Floating Point Operations (FLOPs) and the number of trainable parameters. FLOPs quantify the number of operations necessary to execute a single instance on a deep learning model \cite{app12125902}.
For our pipeline, the total FLOPs count and the number of trainable parameters are recorded as 8,596,611 and 8,593,949, respectively.
\section{Conclusion}\label{sec5}
In this paper, we introduced the first public datasets for Persian font recognition to address the limitations of previous datasets. Alongside these datasets, we propose a pipeline based on CNN models for Persian font recognition. Notably, these neural network types have not been employed in recent papers.
Our method leverages the CNN models' ability to be content- and background-independent, eliminating the need for handcrafted features such as Gabor features, which have been utilized in recent studies.

The experimental results demonstrate that our proposed pipeline achieves a top-1 accuracy of 78.0\% on our new datasets, 89.1\% on the IDPL-PFOD dataset, and 94.5\% on the KAFD dataset. Moreover, the average processing time for a single sample in our proposed datasets is 0.54 seconds for CPU and 0.017 seconds for GPU. For future research, we recommend exploring the design of new CNN architectures or incorporating recently proposed CNN blocks from recent papers.

\bibliographystyle{IEEEtran}
\bibliography{references}

\begin{thebibliography}{10}
\providecommand{\url}[1]{#1}
\csname url@samestyle\endcsname
\providecommand{\newblock}{\relax}
\providecommand{\bibinfo}[2]{#2}
\providecommand{\BIBentrySTDinterwordspacing}{\spaceskip=0pt\relax}
\providecommand{\BIBentryALTinterwordstretchfactor}{4}
\providecommand{\BIBentryALTinterwordspacing}{\spaceskip=\fontdimen2\font plus
\BIBentryALTinterwordstretchfactor\fontdimen3\font minus
  \fontdimen4\font\relax}
\providecommand{\BIBforeignlanguage}[2]{{%
\expandafter\ifx\csname l@#1\endcsname\relax
\typeout{** WARNING: IEEEtran.bst: No hyphenation pattern has been}%
\typeout{** loaded for the language `#1'. Using the pattern for}%
\typeout{** the default language instead.}%
\else
\language=\csname l@#1\endcsname
\fi
#2}}
\providecommand{\BIBdecl}{\relax}
\BIBdecl

\bibitem{23}
\BIBentryALTinterwordspacing
W3Techs. World wide web technology surveys. Accessed Mar. 12, 2022. [Online].
  Available: \url{https://w3techs.com}
\BIBentrySTDinterwordspacing

\bibitem{31}
T.~C. Wei, U.~U. Sheikh, and A.~A.-H.~A. Rahman, ``Improved optical character
  recognition with deep neural network,'' in \emph{2018 IEEE 14th International
  Colloquium on Signal Processing \& Its Applications (CSPA)}, 2018, pp.
  245--249.

\bibitem{26}
Y.~Zhu, T.~Tan, and Y.~Wang, ``Font recognition based on global texture
  analysis,'' \emph{IEEE Transactions on Pattern Analysis and Machine
  Intelligence}, vol.~23, no.~10, pp. 1192--1200, 2001.

\bibitem{27}
S.~Huang, Z.~Zhong, L.~Jin, S.~Zhang, and H.~Wang, ``Dropregion training of
  inception font network for high-performance chinese font recognition,''
  \emph{Pattern Recognition}, vol.~77, pp. 395--411, 2018.

\bibitem{24}
H.~Luqman, S.~A. Mahmoud, and S.~Awaida, ``Arabic and farsi font recognition:
  Survey,'' \emph{International Journal of Pattern Recognition and Artificial
  Intelligence}, vol.~29, no.~01, pp. 1\,553\,002:1--1\,553\,002:23, 2015.

\bibitem{21}
S.~La~Manna, A.~Colia, and A.~Sperduti, ``Optical font recognition for
  multi-font ocr and document processing,'' in \emph{Proceedings. Tenth
  International Workshop on Database and Expert Systems Applications. DEXA 99},
  1999, pp. 549--553.

\bibitem{25}
G.~Chen, J.~Yang, H.~Jin, J.~Brandt, E.~Shechtman, A.~Agarwala, and T.~X. Han,
  ``Large-scale visual font recognition,'' in \emph{Proceedings of the IEEE
  Conference on Computer Vision and Pattern Recognition (CVPR)}, June 2014.

\bibitem{22}
Z.~Wang, J.~Yang, H.~Jin, E.~Shechtman, A.~Agarwala, J.~Brandt, and T.~S.
  Huang, ``Deepfont: Identify your font from an image,'' in \emph{Proceedings
  of the 23rd ACM International Conference on Multimedia}.\hskip 1em plus 0.5em
  minus 0.4em\relax New York, NY, USA: Association for Computing Machinery,
  2015, p. 451–459.

\bibitem{34}
S.~Izadi, M.~Haji, and C.~Y. Suen, ``A new segmentation algorithm for online
  handwritten word recognition in persian script,'' in \emph{Proc. Eleventh
  International Conf. Frontiers in Handwriting Recognition (CFHR 2008)}.\hskip
  1em plus 0.5em minus 0.4em\relax Citeseer, 2008, pp. 598--603.

\bibitem{16}
\BIBentryALTinterwordspacing
A.~Keipour, M.~Eshghi, S.~M. Ghadikolaei, N.~Mohammadi, and S.~Ensafi,
  ``Omnifont persian ocr system using primitives,'' \emph{ArXiv}, 2022.
  [Online]. Available: \url{https://arxiv.org/abs/2202.06371}
\BIBentrySTDinterwordspacing

\bibitem{36}
S.~Pouyanfar, S.~Sadiq, Y.~Yan, H.~Tian, Y.~Tao, M.~P. Reyes, M.-L. Shyu, S.-C.
  Chen, and S.~S. Iyengar, ``A survey on deep learning: Algorithms, techniques,
  and applications,'' \emph{ACM Comput. Surv.}, vol.~51, no.~5, sep 2018.

\bibitem{32}
D.~Gabor, ``Theory of communication. part 1: The analysis of information,''
  \emph{Journal of the Institution of Electrical Engineers-Part III: Radio and
  Communication Engineering}, vol.~93, no.~26, pp. 429--441, 1946.

\bibitem{7}
M.~B. Imani, M.~R. Keyvanpour, and R.~Azmi, ``Semi-supervised persian font
  recognition,'' \emph{Procedia Computer Science}, vol.~3, pp. 336--342, 2011.

\bibitem{11}
K.~Eghbali, H.~Veisi, M.~Mirzaie, and Y.~M. Behbahani, ``Font recognition for
  persian optical character recognition system,'' in \emph{2017 10th Iranian
  Conference on Machine Vision and Image Processing (MVIP)}.\hskip 1em plus
  0.5em minus 0.4em\relax IEEE, 2017, pp. 252--257.

\bibitem{41}
\BIBentryALTinterwordspacing
M.~Mohammadian. (2022, Jul) Official github repository, persis: A persian font
  recognition pipeline using convolutional neural networks. [Online].
  Available: \url{http://github.com/mehrdad-dev/persis}
\BIBentrySTDinterwordspacing

\bibitem{28}
Y.~LeCun, Y.~Bengio, and G.~Hinton, ``Deep learning,'' \emph{nature}, vol. 521,
  no. 7553, pp. 436--444, 2015.

\bibitem{8}
E.~M. Senobari and H.~Khosravi, ``Farsi font recognition based on combination
  of wavelet transform and sobel-robert operator features,'' in \emph{2012 2nd
  International eConference on Computer and Knowledge Engineering
  (ICCKE)}.\hskip 1em plus 0.5em minus 0.4em\relax IEEE, 2012, pp. 29--33.

\bibitem{1}
A.~Borji and M.~Hamidi, ``Support vector machine for persian font
  recognition,'' \emph{International Journal of Computer Systems Science and
  Engineering}, vol.~2, no.~3, 2007.

\bibitem{2}
H.~Khosravi and E.~Kabir, ``Farsi font recognition based on sobel--roberts
  features,'' \emph{Pattern Recognition Letters}, vol.~31, no.~1, pp. 75--82,
  2010.

\bibitem{38}
W.~Khan, ``Image segmentation techniques: A survey,'' \emph{Journal of Image
  and Graphics}, vol.~1, no.~4, pp. 166--170, 2013.

\bibitem{33}
S.~Minaee, Y.~Boykov, F.~Porikli, A.~Plaza, N.~Kehtarnavaz, and D.~Terzopoulos,
  ``Image segmentation using deep learning: A survey,'' \emph{IEEE Transactions
  on Pattern Analysis and Machine Intelligence}, vol.~44, no.~7, pp.
  3523--3542, 2022.

\bibitem{15}
O.~Ronneberger, P.~Fischer, and T.~Brox, ``U-net: Convolutional networks for
  biomedical image segmentation,'' in \emph{Medical Image Computing and
  Computer-Assisted Intervention -- MICCAI 2015}.\hskip 1em plus 0.5em minus
  0.4em\relax Springer International Publishing, 2015, pp. 234--241.

\bibitem{37}
\BIBentryALTinterwordspacing
J.~C. Ye and W.~K. Sung, ``Understanding geometry of encoder-decoder cnns,''
  2019. [Online]. Available: \url{https://arxiv.org/abs/1901.07647}
\BIBentrySTDinterwordspacing

\bibitem{43}
K.~Xia, J.~Huang, and H.~Wang, ``Lstm-cnn architecture for human activity
  recognition,'' \emph{IEEE Access}, vol.~8, pp. 56\,855--56\,866, 2020.

\bibitem{19}
A.~Krizhevsky, I.~Sutskever, and G.~E. Hinton, ``Imagenet classification with
  deep convolutional neural networks,'' in \emph{Advances in Neural Information
  Processing Systems}, F.~Pereira, C.~J.~C. Burges, L.~Bottou, and K.~Q.
  Weinberger, Eds., vol.~25.\hskip 1em plus 0.5em minus 0.4em\relax Curran
  Associates, Inc., 2012.

\bibitem{42}
R.~Ramanathan, K.~Soman, L.~Thaneshwaran, V.~Viknesh, T.~Arunkumar, and
  P.~Yuvaraj, ``A novel technique for english font recognition using support
  vector machines,'' in \emph{2009 International Conference on Advances in
  Recent Technologies in Communication and Computing}, 2009, pp. 766--769.

\bibitem{10}
M.~Ziaratban and F.~Bagheri, ``Improving farsi font recognition accuracy by
  using proposed directional elliptic gabor filters,'' in \emph{2013 First
  Iranian Conference on Pattern Recognition and Image Analysis (PRIA)}.\hskip
  1em plus 0.5em minus 0.4em\relax IEEE, 2013, pp. 1--5.

\bibitem{3}
A.~A. Hajiannezhad and S.~Mozaffari, ``Fractal and multi-fractal dimensions for
  farsi/arabic font type and size recognition,'' in \emph{2011 7th Iranian
  Conference on Machine Vision and Image Processing}.\hskip 1em plus 0.5em
  minus 0.4em\relax IEEE, 2011, pp. 1--4.

\bibitem{5}
M.~Zahedi and S.~Eslami, ``Farsi/arabic optical font recognition using sift
  features,'' \emph{Procedia Computer Science}, vol.~3, pp. 1055--1059, 2011.

\bibitem{12}
Z.~Hossein-Nejad, H.~Agahi, and A.~Mahmoodzadeh, ``Farsi font detection using
  the adaptive rkem-surf algorithm,'' \emph{Information Systems \&
  Telecommunication}, p. 188, 2020.

\bibitem{6}
Y.~Pourasad, H.~Hassibi, and A.~Ghorbani, ``Farsi font face recognition in
  letter level,'' \emph{Procedia Technology}, vol.~1, pp. 378--384, 2012.

\bibitem{4}
Y.~Pourasad, H.~Hassibi, and A.~Ghorbani, ``Farsi font recognition using holes
  of letters and horizontal projection profile,'' in \emph{International
  Conference on Innovative Computing Technology}.\hskip 1em plus 0.5em minus
  0.4em\relax Springer, 2011, pp. 235--243.

\bibitem{9}
M.~Ziaratban and F.~Bagheri, ``Farsi font recognition based on the fonts of
  text samples extracted by som,'' \emph{Journal of Mathematics and Computer
  Science}, vol.~15, pp. 40--56, 07 2015.

\bibitem{30}
\BIBentryALTinterwordspacing
Unsplash. The official unsplash api. Accessed Mar. 15, 2022. [Online].
  Available: \url{https://unsplash.com/developers}
\BIBentrySTDinterwordspacing

\bibitem{13}
\BIBentryALTinterwordspacing
Dataheart. Persian text of the shahnameh book. Accessed Mar. 12, 2022.
  [Online]. Available: \url{http://dataheart.ir}
\BIBentrySTDinterwordspacing

\bibitem{14}
\BIBentryALTinterwordspacing
Bigdata. Persian dictionary. Accessed Mar. 12, 2022. [Online]. Available:
  \url{https://bigdata-ir.com/}
\BIBentrySTDinterwordspacing

\bibitem{17}
D.~Kingma and J.~Ba, ``Adam: A method for stochastic optimization,''
  \emph{International Conference on Learning Representations}, 12 2014.

\bibitem{18}
\BIBentryALTinterwordspacing
M.~Lin, Q.~Chen, and S.~Yan, ``Network in network,'' 2013. [Online]. Available:
  \url{https://arxiv.org/abs/1312.4400}
\BIBentrySTDinterwordspacing

\bibitem{20}
H.~Luqman, S.~A. Mahmoud, and S.~Awaida, ``Kafd arabic font database,''
  \emph{Pattern Recognition}, vol.~47, no.~6, pp. 2231--2240, 2014.

\bibitem{29}
F.~s. Hosseini, S.~Kashef, E.~Shabaninia, and H.~Nezamabadi-pour, ``Idpl-pfod:
  An image dataset of printed farsi text for ocr research,'' in
  \emph{Proceedings of The Second International Workshop on NLP Solutions for
  Under Resourced Languages (NSURL 2021) co-located with ICNLSP 2021}.\hskip
  1em plus 0.5em minus 0.4em\relax Trento, Italy: Association for Computational
  Linguistics, 12--13 Nov. 2021, pp. 22--31.

\bibitem{app12125902}
M.~Shayestegan, J.~Kohout, K.~Štícha, and J.~Mareš, ``Advanced analysis of
  3d kinect data: Supervised classification of facial nerve function via
  parallel convolutional neural networks,'' \emph{Applied Sciences}, vol.~12,
  no.~12, 2022.

\end{thebibliography}

\end{document}